\documentclass[journal]{IEEEtran}

\usepackage{cite}
\usepackage{orcidlink}
\usepackage{pict2e}
\usepackage{graphicx}
\usepackage{pstricks}
\usepackage{verbatim}
\usepackage{psfrag}
\usepackage[english]{babel}
 \usepackage{amsmath}
 \usepackage{amssymb}
 \usepackage[font=small]{caption}
  \usepackage{fancybox}
 \usepackage{steinmetz}
\usepackage{subcaption}
\usepackage{bodegraph}
\usepackage{gensymb}
\usepackage{algorithm,algorithmic}
\usepackage{amsfonts}
\usepackage{dsfont}
\usepackage{siunitx}
\usepackage{lscape}
\usepackage{enumerate}
\usepackage{lmodern}
\usepackage[T1]{fontenc}
\usepackage[framed]{matlab-prettifier}
\usepackage{mathtools}
\usepackage{color} 
\usepackage{xcolor,colortbl}
\usepackage{colortbl}
\usepackage{multirow, hhline}
\usepackage{url}
\usepackage{hyperref}
\usepackage{multicol}
\usepackage{supertabular}

  \def\c{{\boldsymbol c}} 
\def\e{{\boldsymbol e}} \def\f{{\boldsymbol f}} \def\g{{\boldsymbol g}} 
   
\def\m{{\boldsymbol m}} \def\n{{\boldsymbol n}}  \def\p{{\boldsymbol p}}
 \def\r{{\boldsymbol r}} \def\s{{\boldsymbol s}} 
\def\u{{\boldsymbol u}}   \def\x{{\boldsymbol x}}
\def\y{{\boldsymbol y}} 

\def\A{{\boldsymbol A}} \def\B{{\boldsymbol B}}  
 \def\F{{\boldsymbol F}}  \def\H{{\boldsymbol H}}
 \def\J{{\boldsymbol J}} \def\K{{\boldsymbol K}} 
\def\M{{\boldsymbol M}}   
\def\Q{{\boldsymbol Q}} \def\R{{\boldsymbol R}} \def\S{{\boldsymbol S}}

\newtheorem{Remar}{\bf Remark}

\DeclareMathOperator*{\argmin}{argmin} 
\def\m{\mbox{$\boldsymbol{\mu}$}} 

\definecolor{orangeG}{cmyk}{0 75 100 0}
\definecolor{greenG}{cmyk}{75 0 100 0}

\begin{document}
\title{A Minimum-Jerk Approach to Handle Singularities in Virtual Fixtures}
\author{Giovanni Braglia$^{1}$\orcidlink{0000-0002-2230-8191},Sylvain Calinon$^{2}$\orcidlink{0000-0002-9036-6799}, Luigi Biagiotti$^{1}$\orcidlink{0000-0002-2343-6929}
\thanks{*This work has been partially funded by the ECOSISTER Project. Project funded under the National Recovery and Resilience Plan (NRRP), Mission 04 Component 2 Investment 1.5 – NextGenerationEU, Call for tender n. 3277 dated 30/12/2021. Award Number:  0001052 dated 23/06/2022.}
\thanks{$^{1}$The authors are with the engineering department Enzo Ferrari of the University of Modena and Reggio Emilia, Italy
{\tt\small \{giovanni.braglia, luigi.biagiotti\}@unimore.it}.\\
$^{2}$ The author is with the Idiap Research Institute, Martigny, Switzerland and the Ecole Polytechnique Fédérale de Lausanne (EPFL), Switzerland. {\tt\small sylvain.calinon@idiap.ch}}%
}
%


\maketitle

\begin{abstract}

Implementing virtual fixtures in guiding tasks constrains the movement of the robot's end effector to specific curves within its workspace. However, incorporating guiding frameworks may encounter discontinuities when optimizing the reference target position to the nearest point relative to the current robot position.
This article aims to give a geometric interpretation of such discontinuities, with specific reference to the commonly adopted Gauss-Newton algorithm. The effect of such discontinuities, defined as Euclidean Distance Singularities, is experimentally proved.
We then propose a solution that is based on a linear quadratic tracking problem with minimum jerk command, then compare and validate the performances of the proposed framework in two different human-robot interaction scenarios.
\end{abstract}


\IEEEpeerreviewmaketitle

\section{Introduction}\label{sec:intro}

\IEEEPARstart{A}{s} collaborative robotics research is making great strides towards safe Human-Robot Interaction (HRI), users are becoming more confident to utilize robots in co-manipulation tasks. 
In such scenarios, it is common to adopt Virtual Fixtures (VFs) to constraint the robot's motion to a specific manifold~\cite{rosenberg1993use}. Especially in those situations where the human is required to manually execute precise tasks, it has been proven that the use of VFs facilitates adherence to the desired task while reducing the mental workload required to maintain accuracy~\cite{selvaggio2018passive,bowyer2013active}.

\begin{figure}[t]
    \centering
    \psfrag{a}[t][t][0.8]{\textcolor{orangeG}{\hspace{-10pt}$\y_0$}$\leftrightarrow\!\m($\textcolor{orangeG}{$s_0$}$)$}
    \psfrag{b}[t][t][0.8]{\textcolor{greenG}{\hspace{20pt}$\x_0$}$\leftrightarrow\!\m($\textcolor{greenG}{$s_0$}$)$}
    \psfrag{c}[t][t][0.8]{\textcolor{orangeG}{\hspace{20pt}$\y_1$}$\leftrightarrow\!\m($\textcolor{orangeG}{$s_1$}$)$}
    \psfrag{d}[t][t][0.8]{$\Delta$}
    \psfrag{e}[t][t][0.8]{\textcolor{greenG}{\hspace{-15pt}$\x_1$}$\leftrightarrow\!\m($\textcolor{greenG}{$s_1$}$)$}
    \psfrag{f}[t][t][0.8]{\textcolor{orangeG}{$\hspace{10pt}\y_2$}$\leftrightarrow\!\m($\textcolor{orangeG}{?}$)$}
    \psfrag{g}[t][t][0.8]{\textcolor{greenG}{\hspace{20pt}$\x_2$}$\leftrightarrow\!\m($\textcolor{greenG}{$s_2$}$)$}
    \psfrag{h}[t][t][0.8]{\textcolor{greenG}{$\x_3$}$\leftrightarrow\!\m($\textcolor{greenG}{$s_3$}$)\; \big( =\m($\textcolor{greenG}{$s_2$}$) \big)$}
    \psfrag{i}[t][t][0.8]{\textcolor{greenG}{\hspace{20pt}$\x_4$}$\leftrightarrow\!\m($\textcolor{greenG}{$s_4$}$)$}
    \psfrag{l}[t][t][1.5]{}

    \includegraphics[trim={0 2cm 0 0},width=0.7\linewidth,width=0.7\linewidth]{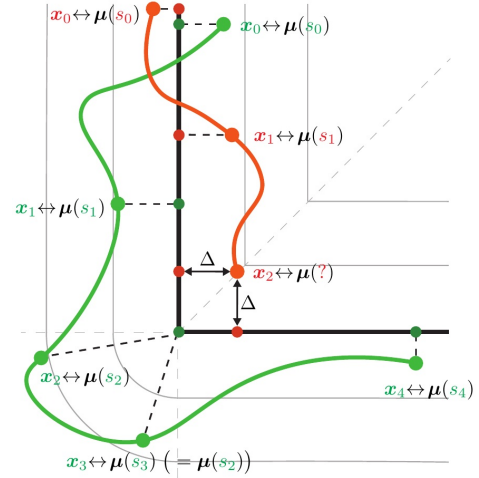}
    \vspace{3mm}
    \caption{ 2D visualization of an Euclidean distance singularity. Continuous gray lines represent the iso-lines with respect to the constraint path $\m(s)$ depicted in black. While a distance-based method could correctly update the reference position $\m(s_t)$ for the green trajectory, it fails to find a stable solution for the red trajectory, as $\y_2$ has the same distance $\Delta$ from 2 points in $\m(s)$.} 
\label{fig:sdf_prob}
\end{figure}

In many guiding applications, allowing compliance with respect to the VF curve $\m$ ensures a more natural and intuitive interaction with the user~\cite{raiola2015co,amirabdollahian2002minimum, sciavicco}. A common approach in such cases, is to adjust the reference position on $\m$ to the point that minimizes the displacement with the actual position of the robot. In this way, the robot is able to follow the user movements, whilst imposing the virtual constraint~\cite{selvaggio2018passive}.
As far as we know, we noticed that this procedure has not yet been extended to positions in the workspace that share the same distance to multiple points on the VF, here called \textit{Euclidean distance singularities} (EDSs). This problem is illustrated in Fig.~\ref{fig:sdf_prob} and investigated throughout this article.

\subsection{Related Works}\label{subsec:related_works}

The term virtual fixture (VF) is used when restricting the robot movements to a specified manifold, and it owes much of its popularity to guiding applications~\cite{rosenberg1993use,bowyer2013active}.
Here, the human interacts with the environment while being enforced by the robot to adhere to a specific curve~\cite{raiola2015co}. 
%
%
Usually, the definition of VFs restricts to the imposition of geometrical constraints not associated with any time-law. When focusing on path planning, it is common practice to define geometrical and time constraints in separates moment~\cite{sciavicco,verscheure2009time}. Given a geometrical constraint $\m$, the time-law can be managed through the definition of the so-called phase variable $s_t$ to further define the dynamics $\m(s_t)$. In this way, one can properly compute $s_t$ to regulate the manipulator dynamics depending on geometrical, kinematic and external constraints~\cite{bianco2017scaling,braglia2023online}.
To the best of our knowledge, we believe that the methods for determining the time-law can be categorized into three main groups: linear, user-defined and optimization-based.

In linear approaches, the time scaling is adjusted by proportionally modify the duration of the task execution $T$, typically varying a scalar value to regulate the velocity of the phase variable $\dot s_t$. 
A classic example of this category is Dynamic Movement Primitives (DMPs), where a nonlinear term controlling the system's dynamics is regulated by a tunable phase variable, provided by the canonical system~\cite{ijspeert2013dynamical}. Another example is provided by~\cite{braglia2023online}, where a scalar parameter adjusts online the velocity of the manipulator to slow down when approaching the human. 

Depending on the type of application, one may want to influence the variation of $s_t$ in a different way. We refer to user-defined approaches whenever the definition of the time-law is made explicit through an analytic formulation that is not scalar on $\dot s_t$. Again, some variations of DMPs transform the canonical system such that $\dot s_t$, for example, depends on the position error~\cite{ijspeert2002movement} or external force measurements~\cite{sidiropoulos2021reversible,braglia2024phase}. Another example can be found in~\cite{raiola2015co}, where authors utilize a dynamical system representation for the computation of $\dot s_t$ along the virtual constraint.

The last category, which will be the focus of this article, includes all the techniques where the time-law $s_t$ is obtained as the solution of an optimization problem~\cite{reynoso2013time,verscheure2009time}. In particular, we observed that many virtual fixture applications rely on distance-based methods, selecting the point on the curve $\m(s)$ that minimizes the Euclidean distance from the robot's end effector (EE) position $\x$~\cite{girgin2023projection,selvaggio2018passive,bischof2016combined,zhang2020assist}. Finding an analytical solution to calculate the minimum distance point $\m(s^*)$ is generally non-trivial. Nevertheless, the presence of techniques such as the Gauss-Newton (GN) algorithm, simplifies the problem to the computation of the phase $s_t^*$ that minimizes the residual $||\x_t - \m(s_t^*)||$~\cite{wang2012gauss}.
The GN algorithm is known to be computationally efficient and was applied in various robotic applications~\cite{khatib2022constraint,rcfs}. However, its use to update $s_t$ may not lead to a convex formulation for the minimum distance problem~\cite{luenberger1984linear}. In this article,
we refer to this condition as \textit{Euclidean-distance singularity} (EDS). We demonstrate analytically and experimentally that approaching an EDS might produce abrupt changes in the phase velocity $\dot s_t$.


Controlling the phase velocity is common in planning algorithms, and the outcome solution usually produce a time-optimal trajectory that exploits the maximum accelerations/torques given the constraints of the robot actuators~\cite{verscheure2009time}. This could potentially result in acceleration profiles with a high rate of change, which can stress the actuators leading to long term damage~\cite{palleschi2019time}.
In these cases, enhancing smoothness is a good compromise between minimum time task execution, and acceleration noise reduction. To achieve so, it is a common practice to minimize the jerk of the planned trajectories~\cite{piazzi2000global,bianco2017scaling}.
Minimum jerk trajectories have been widely investigated also in the reproduction of human arm movements~\cite{biess2007computational,sharkawy2021minimum}. As demonstrated in~\cite{todorov1998smoothness}, minimum jerk trajectories can faithfully reproduce the speed profile of arm movements in reaching tasks or curve tracking, making them suitable for human-robot interaction scenarios~\cite{amirabdollahian2002minimum}.
This motivates our research to propose a novel methodology for the online phase update in guiding VFs, exploiting a jerk-controlled framework based on minimum distance, while ensuring robustness against EDSs.

\subsection{Contribution}\label{subsec:contribution}

The main contributions of this paper are:
(1) present a theoretical representation of the so-called Euclidean distance singularities (EDSs) with an experimental proof of their limitation; (2) provide a formulation for updating the phase variable using minimum jerk control with a linear quadratic tracking (LQT) algorithm and (3), a real-time application of the proposed framework in a human-robot interaction setup.
The outline of this paper is as follows.
Section~\ref{sec:background}  offers a theoretical background on the addressed EDSs problem. Section~\ref{sec:methodology} presents a jerk-based LQT formulation for optimal phase update. Section~\ref{sec:experiments_results} validates and discusses the implementation of our proposed methodology in two experimental scenarios, and Section~\ref{sec:conclusion} concludes this work. 

\section{Background}\label{sec:background} 

In this article, we aim to constrain the robot to follow a predefined reference path $\m$ in the task space. Once established, the operator can use the robot's guidance to navigate along this path. Details of this mechanism are provided in the following paragraphs.

\subsection{Definition of the virtual constrain}\label{subsec:virtual_constrain}

To define the constraint $\m$, we first kinesthetically move the robot end effector to demonstrate the target curve. The recordings are then filtered using the spatial sampling (SS) algorithm from~\cite{braglia2024phase}, to extract the geometrical path information regardless of the timing introduced during the demonstration. This associates the recordings $\r(t) = [r_{x}(t), ,r_{y}(t),, r_{z}(t)]^\top$ with their filtered counterparts $\r_{\Delta,k} = [r_{x,\Delta,k}, ,r_{y,\Delta,k},, r_{z,\Delta,k}]^\top$ and curvilinear coordinates $s_k \in [0,L]$, with $k=1,...,M$ and $L=\Delta M$ representing the curve length $\r_{\Delta}$.
The free parameter $\Delta$ defines the geometric distance between consecutive samples in $\r_{\Delta,k}$ and is such that, given an analytical approximation $\m(s)\approx\r_{\Delta,k}$, the following property holds:
\begin{equation}\label{eq:unitary_norm} 
{\left\lVert\dfrac{d \m(s)}{d s}\right\rVert}_{s=s_k} \approx
\dfrac{\|\r_{\Delta,k+1}-\r_{\Delta,k}\|}{\|s_{k+1}-s_{k}\|}
=\!\dfrac{\Delta}{\Delta}\!
=1.
\end{equation}
We expressed $\m(s)$ using basis functions approximation, representing the signal $\r_{\Delta,k}$ as a weighted sum of nonlinear terms, namely~\cite{calinon2020mixture}
\begin{equation}\label{eq:basisfun}
   \m( \!s_t\!)  \! = \! \! \sum_{i=1}^N \omega_i \phi_i(s_t)  \!= \! \boldsymbol{\omega}^\top  \! \boldsymbol{\phi}(s_t).
\end{equation}
In~\eqref{eq:basisfun}, the constraint curve is formed by summing $N$ basis functions $\phi_i$, each weighted by $\omega_i$, typically found via a Least-Squares solution~\cite{ijspeert2013dynamical}, i.e.
\begin{equation}\label{eq:regression}
    \boldsymbol{\omega} = \arg \! \min_{\!\!\!\!\!\!\!\!\!\!\!\boldsymbol{\omega}} \sum_{k=0}^{M} ||\r_{\Delta,k} - \boldsymbol{\omega}^\top\boldsymbol{\phi}(s_k)  ||^2,
\end{equation}
indicating with $||\cdot||$ the Euclidean norm of the residual term.
The basis functions $\phi_i(s_t)$ can be chosen in different ways, in this case we used Bernstein polynomials~\cite{biagiotti2008trajectory,calinon2020mixture}. Again, the phase variable is denoted by $s_t$, but thanks to the SS algorithm it carries an additional meaning of being the curvilinear coordinate of the curve $\m$.

For instance, with the expression in \eqref{eq:basisfun} we can compute the velocity and acceleration of $\m(s_t)$ respectively as
\begin{equation}\label{eq:velocity_acceleration}
\begin{array}{c}
    \dot{\m}(s_t) = \left.\dfrac{\partial \m(s(t))}{\partial t}\right|_{t=\alpha\Delta t} = \m^{\prime}(s_t)\dot{s}_t \;\,\, \mbox{and}
     \\[4mm]
     \ddot{\m}(s_t)  = \left.\dfrac{\partial^2 \m(s(t))}{\partial t^2}\right|_{t=\alpha\Delta t}= 
     \m^{\prime\prime}(s_t)\dot{s}_t^2 + \m^{\prime}(s_t)\ddot{s}_t,
\end{array}
\end{equation}
with $\ddot{s}_t$ and $\dot{s}_t$ obtained from their respective continuous counter part at sampling time $\Delta t$.
In \eqref{eq:velocity_acceleration} and throughout this article, the derivative with respect to $s_t$ is denoted with the prime symbol, while the discrete time derivative uses the dot notation. Note that, because of~\eqref{eq:unitary_norm}, $\m^{\prime}(s_t)$ will always be well-defined and different from zero~\cite{sciavicco,biagiotti2008trajectory}.

\subsection{Tracking Problem Statement}\label{subsec:problem_statement}

Let $\m(s) \in \mathbb{R}^3$ be the path constraint parametrized with respect to the curvilinear coordinate $s$, and  $\x_t \in \mathbb{R}^3$ be the Cartesian position of the robot's end effector (EE). 
The objective of limiting the EE movement along the curve $\m(s)$ can be formulated as a tracking problem, that is
\begin{equation}\label{eq:optimal_problem}
    s_t^\star = \argmin_{s_t \in [0, M\Delta] }  \| \x_t - \m(s_t) \|^2.  
\end{equation}
The optimal problem in \eqref{eq:optimal_problem} equals to update $s_t$ at every instant $t$ in order to find the closest point $\m(s_t^\star)$ to $\x_t$.
From ~\cite{luenberger1984linear}, the existence of a solution is guaranteed if:
\begin{subequations}\label{eq:optimal_problem_conditions}
    \begin{equation}\label{eq:first_cond_opti}
     \big( \x_t - \m(s_t^\star) \big)^\top \m^{\prime}(s_t^\star) = 0, \ \ \mbox{and} 
    \end{equation}
    \begin{equation}\label{eq:second_cond_opti}
    \| \m^{\prime}(s_t^\star) \|^2 - \big( \x_t - \m(s_t^\star) \big)^\top \m^{\prime\prime}(s_t^\star) > 0.
    \end{equation}
\end{subequations}
The conditions in \eqref{eq:optimal_problem_conditions} leads to the following geometrical intuition. On one hand, the first order necessary condition in \eqref{eq:first_cond_opti} states that the optimal $s_t^\star$ lies on the point $\m(s_t^\star)$ whose tangential component $\m^{\prime}(s_t^\star)$ is orthogonal to the EE position $\x_t$. From~\eqref{eq:unitary_norm}, $\m^{\prime}(s_t)$ represents the tangential unit vector to the curve. 
On the other hand, the second-order sufficient condition in \eqref{eq:second_cond_opti} guarantees the convexity of \eqref{eq:optimal_problem} if the Hessian of the cost function $ \| \x_t - \m(s_t) \|^2$ is positive definite. In particular, given the constraint $\m(s)$ for $s\in[0, \Delta M]$, at every instant $t$ a feasible subset $\boldsymbol{\chi}_t = \big\{ \bar\x_t \in \mathbb{R}^3 : \mbox{~\eqref{eq:second_cond_opti} holds} \big\}$ can be deduced~\cite{bischof2016combined}.
Equation~\eqref{eq:optimal_problem_conditions} implies $\m \in C^2$, which can be ensured by a proper choice of the basis functions in~\eqref{eq:basisfun}~\cite{calinon2020mixture}.

\subsection{Gauss-Newton algorithm}\label{subsec:gauss_newton}

One common technique for solution of the optimization problem in~\eqref{eq:optimal_problem} comes from the Gauss-Newton (GN) algorithm~\cite{wang2012gauss,rcfs}.
In this case, the evolution of the curvilinear coordinate $s_t$ is given by $ s_{t+1} = s_t + \Delta s$, with the term $\Delta s$ being the GN update.
Defining the residual $\f(s_t)=\x_t - \m(s_t)$, the GN algorithm provides an analytical solution for the computation of $\Delta s$, that is
\begin{equation}\label{eq:delta_s}
\begin{array}{c}
    \Delta s_t = - \H ( s_t )^{-1} \g (s_t), \ \ \mbox{with} \\[2
    mm]
    \g(s_t) = 2 \J_{\f}^\top \f(s_t), \ \ \H(s_t) \approx  2 \J_{\f}^\top \J_{\f} .
\end{array}
\end{equation}
In~\eqref{eq:delta_s}, $\J_{\f} \in  \mathbb{R}^{3\times D}$ defines the Jacobian matrix for $\f(s_t) \in \mathbb{R}^3$, while $\g \in \mathbb{R}^D$ and $\H \in \mathbb{R}^{D\times D}$ are the gradient vector and the Hessian matrix of the cost function with respect to the residual $\f(s_t)$, respectively. For our tracking problem we can assume $\J_{\f}(s_t) = -\m^{\prime}(s_t)$.
The advantage of the GN algorithm is that, despite being a second order technique, it converges in few steps. This is possible as the Hessian can be easily estimated from the Jacobian in~\eqref{eq:delta_s}, which guarantees its positive definiteness~\cite{wang2012gauss,rcfs}.

Note that the GN algorithm neglects the second-order derivative term of the Hessian in~\eqref{eq:second_cond_opti}. However, when this term cannot be ignored, the GN algorithm may experience slow convergence.
From~\eqref{eq:unitary_norm} one can write~\eqref{eq:second_cond_opti} as
\begin{equation}\label{eq:second_cond_opti_bis}
    1 - \big( \x_t - \m(s_t) \big)^\top \m^{\prime\prime}(s_t) > 0.  
\end{equation}
Given the geometry of curves, we have that $\m^{\prime\prime}(s_t) =\n(s_t)\kappa(s_t)$, where $\n$ and $\kappa$ define the normal versor and the curvature, respectively, calculated at the curvilinear coordinate $s_t$~\cite{toponogov2006differential,sciavicco}. In particular, observing that $r(s_t)=1/\kappa(s_t)$ expresses the radius of the osculating circle at the point $\m(s_t)$, the condition in~\eqref{eq:second_cond_opti_bis} can be written as
\begin{equation}\label{eq:osculating}
    \big( \x_t - \m(s_t) \big)^\top \n(s_t) < r(s_t).  
\end{equation}
Equation~\eqref{eq:osculating} provides the following geometric intuition: given the reference $\m(s_t)$, the center of the osculating sphere defines a boundary for the deviation of the EE position $\x_t$, out of which the second order condition for the optimality in~\eqref{eq:second_cond_opti_bis} is violated.
Therefore, the optimal problem in~\eqref{eq:optimal_problem} no longer admits a unique solution, as it looses convexity due to the fact that the Hessian matrix represented in~\eqref{eq:second_cond_opti} ceases to be positive definite~\cite{luenberger1984linear,rcfs}.

Notably, in~\cite{bischof2016combined} authors demonstrate that the time derivative of the optimal $s_t^\star$ in~\eqref{eq:optimal_problem} can be written as
\begin{equation}\label{eq:opti_phase_vel}
    \dot s_t^\star = \dfrac{\m^{\prime}(s_t^\star)^\top \dot{\x}_t}{1 - \big( \x_t - \m(s_t^\star) \big)^\top \m^{\prime\prime}(s_t^\star)} .
\end{equation}
Given the condition in \eqref{eq:second_cond_opti_bis}, it can be evinced from~\eqref{eq:opti_phase_vel} that the limit case $\big( \x_t - \m(s_t) \big)^\top \m^{\prime\prime}(s_t) \rightarrow 1$ implies $\dot s_t^\star \rightarrow \infty$. Hence, small values of $\dot \x_t$ results in large derivatives of $\dot s_t^\star$ when close to the center of the osculating circle calculated at $\m(s_t^\star)$ . 
In our case, as we allow compliance in the directions outside the path constraint $\m(s)$, this can compromise the contact with the robot and the human, given the generation of the reference trajectories outlined in~\eqref{eq:velocity_acceleration}.
These considerations, that will be further analyzed in Section~\ref{subsec:center_reaching}, provide an analytical explanation of the constraint imposed in~\cite{zhang2020assist} and lead to the definition of Euclidean distance singularity.

%
\begin{Remar}\label{remark1}
\textit{Given the optimal problem in~\eqref{eq:optimal_problem}, with $\bar{s}$ being curvilinear coordinate of the curve $\m(s) \in \mathbf{R}^n$, we denote with Euclidean distance singularity (EDS) the set:}
\begin{equation}\label{eq:eds}    
   EDS = \left\{ \begin{array}{cc}
         \x \in \mathbf{R}^n : & \big( \x - \m(\bar{s}) \big)^\top \m^{\prime}(\bar{s}) = 0,   \\
                               & \big( \x - \m(\bar{s}) \big)^\top \n(\bar{s}) \geq r(\bar{s})
    \end{array} \right\}.
\end{equation}
\end{Remar}

The geometric intuition is that if we move along the normal direction 
$\n(\bar{s})$ by a distance greater than the radius of the osculating circle at $\m(\bar{s})$, the optimization problem in~\eqref{eq:optimal_problem} degenerates into multiple local solutions.
Though these conditions may seem restrictive, Sec.~\ref{subsec:center_reaching} demonstrates that even near an EDS issues can arise.

\begin{figure}[t]

    \centering
\psfrag{F0}[][][0.7]{$\m(s)$}
\psfrag{M0}[t][t][0.8]{$\gamma$}
\psfrag{s0}[t][t][0.8]{$\!\x_{\bar{t}}\!$} 
\psfrag{p1}[t][t][0.8]{$\p_1$}
\psfrag{p2}[t][t][0.8]{$\p_2$}
\psfrag{p3}[t][t][0.8]{$\p_3$}
\psfrag{p4}[t][t][0.8]{$\p_4$}
\psfrag{x}[t][t][0.9]{$x_1$[m]}
\psfrag{y}[t][r][0.9
]{\hspace{3mm}$x_2$[m]}
\psfrag{c1}[t][t][0.8]{$c_{\p_1}$}
\psfrag{c2}[t][t][0.8]{$c_{\p_2}$}
\psfrag{c3}[t][t][0.8]{$c_{\p_3}$}
\psfrag{c4}[t][t][0.8]{$c_{\p_4}$}
\psfrag{c}[t][t][1.0]{cost}
\psfrag{s}[t][t][1.0]{$s$}

    \includegraphics[width=0.9\linewidth]{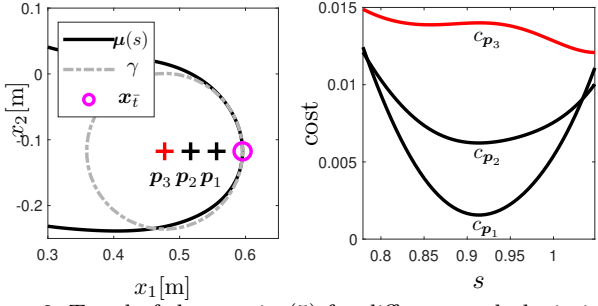}
    \caption{Trend of the cost in \eqref{eq:optimal_problem} for different path deviations.}
\label{fig:cost}
\vspace{-5mm}
\end{figure}

Figure~\ref{fig:cost} provides a 2D example. The left subplot shows the osculating circle $\gamma$ at $\m(s_{\bar{t}})$ in dashed gray. The right subplot depicts the cost functions $c_{\p_1}$, $c_{\p_2}$, and $c_{\p_3}$ in~\eqref{eq:optimal_problem} concerning the EE position $\x_{\bar{t}}$ relative to points $\p_1$, $\p_2$, and $\p_3$. Note that the cost function loses local convexity once the deviation reaches the center of the osculating circle $\p_3$. Note that, for curves with sharp corners, using a basis function approximation as in~\eqref{eq:basisfun} ensures the continuity of the geometrical path and the Jacobian in~\eqref{eq:delta_s}~\cite{biagiotti2008trajectory}.

\section{Methodology}\label{sec:methodology}

With reference to equation~\eqref{eq:opti_phase_vel}, the proximity of the EE position $\x_t$ to EDSs could lead to high velocities $\dot{s}_t$. One way to mitigate such undesired behaviour, is to formulate the optimal problem in~\eqref{eq:optimal_problem} such that we can control the damping on the velocity term. This is possible by considering minimum-jerk trajectories. 
Moreover, many results in literature indicate that this model is well-suited for human-robot co-manipulation tasks~\cite{todorov1998smoothness,amirabdollahian2002minimum,chen2024variable}.

Minimum-jerk approaches highlight that the velocity profile of hand movements is established by minimizing its squared jerk~\cite{biess2007computational}.
Authors in~\cite{meirovitch2016geometrical} extend the minimum jerk formulation to the so-called \textit{jerk-accuracy} (JA) model, where the tracking precision and the control action is modulated with the help of a Lagrange multiplier.
We here formulate the JA model as a linear quadratic tracking (LQT) problem for discrete-time systems. In doing so, the control command is fed to a chain of three discrete-time integrators, i.e. $u_t = \dddot s_t$, such that the reference $\m(s_t)$ stays aligned with the EE position $\x_t$. The problem becomes
\begin{equation}\label{eq:min_jerk}
     \min_{u_t}  \f(\x_t,\s_t)^{\!\top} \Q \f(\x_t,\s_t) + u_t^{\!\top} R u_t,  
\end{equation}
with $\f(\x_t,\s_t) = \big[ \e_t^\top, \dot{\e}_t^\top, \ddot{s}_t\big]^{\!\top}$, $\e_t = \x_t - \m(s_t)$, subject to
\begin{equation}\label{eq:s_dyn}
    \underbrace{\left[ \begin{array}{c}
         s_{t+1}   \\
         \dot{s}_{t+1}   \\
         \ddot{s}_{t+1} 
    \end{array} \right]}_{\s_{t+1}} 
    \!\!=\!\! \underbrace{\left[ \begin{array}{ccc}
         1 & \Delta t & \Delta t^2/2 \\
         0 & 1 & \Delta t \\
         0 & 0 & 1
    \end{array} \right]}_{\A} \!
    \underbrace{\left[\begin{array}{c}
         s_{t}  \\
         \dot{s}_{t}  \\
         \ddot{s}_{t}
    \end{array} \right]}_{\s_t}
    \!\!+\!\! \underbrace{\left[\begin{array}{c}
         0  \\
         0  \\
         \Delta t
    \end{array} \right]}_{\B} \! u_t  .      
\end{equation}
In~\eqref{eq:min_jerk}, $\s_t = [ s_t, \dot{s}_t, \ddot{s}_t ]^\top$, $s_t \in [0, \Delta M]$, $\Delta t$ is the sampling time, $\Q=\mbox{diag}(\c_1, \c_2, c_3)$ is called precision matrix with constant parameters $\c_1,\c_2 \in \mathbb{R}^3$, $c_3 \in \mathbb{R}$, and the constant $R$ is the weight control coefficient. The addition of $\ddot{s}_t$ to the non-linear term $\f$ guarantees that the computation of the Jacobian $\J_t = \partial \f_t / \partial \s_t$ is not rank deficient. Let the robot EE position be at $\bar{\x}_t$. We apply the Newton's method to minimize the cost $c(\s_t,u_t) = \f(\s_t)^{\!\top} \Q \f(\s_t) + u_t^{\!\top} R u_t$. This can be done by carrying out a second order Taylor expansion around the point $(\s_t,u_t)$, that is~\cite{rcfs}:
\begin{equation}\label{eq:taylor_expansion}
    \begin{array}{cc}
       \Delta c( \Delta \s_t, \Delta u_t) \approx  & \hspace{-6mm} 2 \Delta \s_t^{\!\top} \J(\s_t)^{\!\top} \Q \f(\s_t) + 2 \Delta u_t^{\!\top} R u_t + \\ [2mm]
         & + \Delta \s_t^{\!\top} \! \J(\s_t)^{\!\top} \! \Q \J(\s_t) \Delta \s_t + \Delta u_t ^{\!\top} \! R \Delta u_t .
    \end{array}
\end{equation}
Therefore, the optimization problem in~\eqref{eq:min_jerk} is linearized using the cost function in~\eqref{eq:taylor_expansion}, i.e.
\begin{equation}\label{eq:lin_cost}
    \min_{\Delta \u} \Delta c( \Delta \s, \Delta u) \hspace{3 mm} \mbox{s.t.} \hspace{3 mm} \Delta \s = \S_u \Delta \u,
\end{equation}
with $\Delta \s = [\Delta \s_1, ..., \Delta \s_T]^\top$, $\Delta \u = [\Delta u_1, ..., \Delta u_{T-1}]^\top$,  and $\S_u$ characterizing the control transfer matrix of the system's dynamics in~\eqref{eq:s_dyn} expressed at trajectory level for $t=1,2,...,T$, namely $\s = \S_{\s}\s_1 + \S_u \u$~\cite{rcfs}. 
By differentiating~\eqref{eq:lin_cost} with respect to $\Delta \u$ and equating to zero, one can compute at each iteration the optimal control command:
\begin{equation}\label{eq:optimal_control}
    \Delta \u^\star = \big( \S_u^\top \J(\!\s\!)^{\!\top}\! \Q \J(\!\s\!) \S_u \!+\! \R \big)^{\!-1} \big(\! -\!\S_u^\top \J(\!\s\!)^{\!\top}\! \Q \f(\!\s\!)\! -\! \R \u \big).
\end{equation}
With the proposed LQT formulation, the quadratic cost function in~\eqref{eq:min_jerk} offers intuitive and interpretable settings for parameters $\Q$ and $R$. Additionally, a second-order approximation reliably guides the optimization steps towards the minimum of the local estimation of the cost function~\eqref{eq:taylor_expansion}~\cite{luenberger1984linear,rcfs}. While higher-order methods may provide a more accurate approximation, they require additional algorithms to find their minimum, thereby increasing computational effort.

\begin{figure}[t]
    \centering
\psfrag{a}[t][t][1]{$\dot{s}_1$}
\psfrag{b}[t][t][1]{$\dot{s}_2$}
\psfrag{c}[t][t][1]{$\dot{s}_3$}
\psfrag{q}[t][t][1]{$\c_2$}
\psfrag{t}[t][t][1]{$t$[s]}
\psfrag{s}[t][t][1.2]{\hspace{3mm}$\dot s$}
    
    \includegraphics[width=\linewidth
    ]{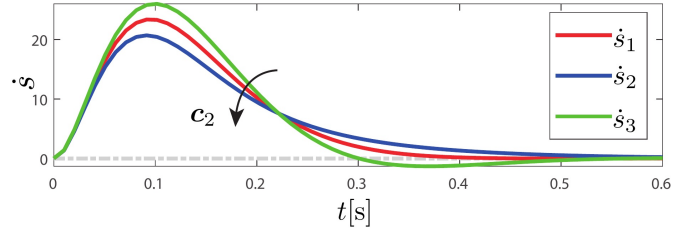}
    \caption{Behaviour of the curvilinear parameter velocity $\dot s$ with varying velocity weight $\c_2$.}
\label{fig:sdot}
\end{figure}

The LQT algorithm described earlier was initially simulated in a reaching task.
The velocity profiles $\dot s$ are displayed in Fig.~\ref{fig:sdot}. The figure demonstrates that the proposed framework achieves the characteristic bell-shaped profile which characterizes human reaching tasks~\cite{todorov1998smoothness}. Moreover, by reducing $\c_2$ in the precision matrix $\Q$, the velocity error $\dot \e = \dot \x_t - \dot \m(s_t)$ becomes less damped, thus the controller more reactive. As a consequence, the velocity profile changes its sign before reaching zero, as for $s_3$ depicted in green. This behaviour is common in rapid arm movements, where a corrective movement is actuated to refine the reaching precision~\cite{meirovitch2016geometrical}.

The implementation of the LQT algorithm is resumed in Algorithm~\ref{algorithm1}.
In its real-time implementation, the LQT algorithm is computed at each iteration with reduced time steps $t=1,2,...,T_W$, using a Model Predictive Control approach. This allows for faster computation of the optimal control $\hat{u}{t+1}$ by minimizing the time window $T_W$. Additionally, the control $\Delta \u^\star$ is iteratively computed until its norm reaches a lower bound $\Delta{min}$, with a maximum number of epochs $I_{M!A!X}$ set to prevent unwanted latency. The initial conditions for each iteration are set to the last commanded pair $(\s_{t-1}, u_{t-1})$, providing a warm start for faster convergence.

\begin{algorithm}[t] 
 \caption{Iterative LQT algorithm}
 \begin{algorithmic}[1]
 \renewcommand{\algorithmicrequire}{\textbf{Input: $\x_{\bar{t}}, \; \m(s_{\bar{t}}), \; \A, \; \B, \; \Q, \; R, $}}
 \renewcommand{\algorithmicensure}{\textbf{Output: $\s_{\bar{t}+1}^\star$}}
 \REQUIRE in~\eqref{eq:min_jerk}-~\eqref{eq:s_dyn}, $ I_{M\!A\!X}$ .
 \ENSURE  .
 \\ \textit{Initialization} : for $t = \bar{t}, \bar{t}\!+\!1, ..., \bar{t}\! + \!T_W \!-\!1 $ compute the transfer matrices $\S_\s, \; \S_\u$ and initialize $\s_{1,0} = \s_{\bar{t}}$, $\u = [ 0_{\bar{t}},...,0_{\bar{t}\!+\!T_W\!-\!1}]$. \\
 
  \FOR {$i = 1$ to $I_{M\!A\!X}$}
  \STATE Compute the dynamics $\s = \S_{\s}\s_{1,i\!-\!1} + \S_u \u$
  \STATE Calculate the residual $\f(\x_{\bar{t}},\s)$ and the Jacobian $\J(\s)$ 
  in~\eqref{eq:min_jerk}-~\eqref{eq:lin_cost}
  \STATE Compute $\Delta \u^\star$ as in~\eqref{eq:optimal_control}
  \IF {($||\Delta \u^\star||< \Delta_{min}$)}
  \STATE Local minimum reached, exit \textbf{for} loop
  \ENDIF
  \STATE Update control $\u = \u + \Delta \u^\star$
  \STATE Update initial state $\s_{1,i}= \A \s_{1,i\!-\!1} + \B u_1 $
  \STATE Re-define the control as $\u = [ u_2, u_3, ..., u_{T_W}, u_{T_W} ]^\top$
  \ENDFOR
  \STATE Save the state $\s_{\bar{t}+1}^\star = \s_{1,i} = [ s_{1,i}, \dot s_{1,i}, \ddot s_{1,i} ]^\top$
 \RETURN  $\s_{\bar{t}+1}^\star$ 
 \end{algorithmic} 
\label{algorithm1}
\end{algorithm}
%


\section{Experiments and Results}\label{sec:experiments_results}

We compare the proposed linear quadratic tracking (LQT) algorithm, together with the Gauss-Newton (GN), a virtual mechanism (VM) approach~\cite{raiola2015co} and a simple control in gravity compensation (GC). The validation consists in two different experimental scenarios which are detailed in the following sections\footnote{An illustrative video of the experiments can be found at:  \textit{https://youtu.be/FLrSDptwb8Q} }.   

\subsection{Experimental Setup}\label{subsec:exp_setup}

\begin{figure}[t]
    \centering
\psfrag{a}[t][t][0.9]{\hspace{2.5mm}\parbox{2.7cm}{\vspace{-1.5mm}\centering Admittance \\Controlled Robot} }
\psfrag{b}[t][t][1.2]{User}
\psfrag{c}[t][t][1.2]{LQT}
\psfrag{d}[t][t][1.2]{VF}
\psfrag{e}[t][t][1.]{\hspace{2mm}$\m(s_t)$}
\psfrag{f}[t][t][1.]{$\F_{t}$}
\psfrag{g}[t][t][1.]{$\x_{t}$}
\psfrag{h}[t][t][1.]{$\s_t$}
\psfrag{i}[t][t][1.]{$\e_t$}
\psfrag{p}[t][t][0.8]{$+$}
\psfrag{m}[t][t][0.8]{$-$}

\includegraphics[width=\linewidth]{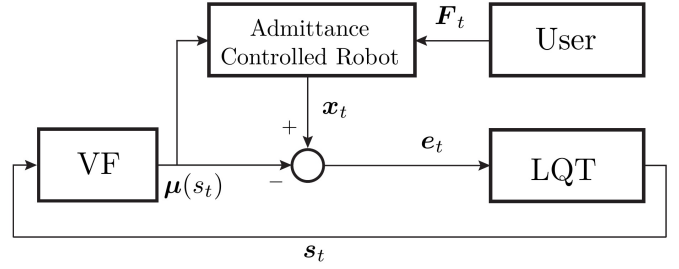}
\caption{Controller framework. Here VF denotes the virtual fixture constraint.}
\label{fig:framework}
\vspace{-5mm}
\end{figure}

Experiments involved a Franka Emika Panda robot together with a Schunk FT-AXIA force/torque sensor. The robot controller has been implemented in C++, while the codes for the optimization of $s_t$ were developed in MATLAB; finally, each node was connected using Robot Operative System (ROS).
The tested algorithms imposes that the generated reference point $\m(s_t)$ follows the end effector (EE) displacement caused by the user~\cite{sciavicco}. To minimize the kinesthetic effort while preventing the user from sensing the natural manipulator's inertia, the robot has been endowed with and admittance control.  
Specifically, the translation dynamics in the Cartesian space have been chosen as:
\begin{equation}\label{eq:simulated_dyn}
    \M \ddot{\hat{\m}}_t + \B \dot{\hat{\m}}_t + \K \big( \hat\m_t - \m(s_t) \big) = \F_{t},
\end{equation}
where $\M=\mbox{diag}(m, m, m)$, $\B=\mbox{diag}(b,b,b)$, $\K=\mbox{diag}(k,k,k)$ are the simulated inertia, damping and stiffness coefficient, respectively, while $\F_{t}$ represents the measured force at the EE.  
To achieve good tracking and low manipulator inertia, we empirically choose $m=1.5 [kg]$, $b=15[N\!\cdot\!s/m]$ and $k=200[N\!\cdot\!m]$. 
The position $\hat\m_t$ computed from the admittance model (\ref{eq:simulated_dyn}) is sent to the robot and used as a reference signal for inverse dynamics position control in Cartesian space~\cite{sciavicco}.
During the experiments, the orientation of the EE was kept fixed. The adopted framework is schematized in Fig.~\ref{fig:framework}.

Finally we consider the values in~\eqref{eq:min_jerk} for control weight $R$ and the precision matrix $\Q=\mbox{diag}(\c_1, \c_2, c_3)$, given $\c_1,\c_2 \in \mathbb{R}^3$ and $c_3 \in \mathbb{R}$. 
The initial values were selected based on the specific path $\m$ needed for the task. Specifically, we employed an inverse LQR approach as in~\cite{priess2014solutions} for computing first attempt values.

%

\subsection{Center-reaching task}\label{subsec:center_reaching}

\begin{figure}
    \centering
\psfrag{t}[t][t][0.8]{$t$[s]}
\psfrag{ mu}[t][t][0.8]{$\m(s)$}
\psfrag{ xgn}[t][t][0.8]{$\x_{\!GN}$}
\psfrag{ xlq}[t][t][0.8]{\hspace{0.5mm}$\x_{\!LQT}$}
\psfrag{ c}[][r][0.8]{$\boldsymbol{\gamma}(\hat{s})$}
\psfrag{x1}[t][t][0.8]{$x$[m]}
\psfrag{x2}[t][t][0.8]{$y$[m]}
\psfrag{m}[t][t][0.9]{\hspace{-7mm}$|\F|$}
\psfrag{a}[t][t][0.9]{\hspace{-7mm}
$\angle\F$}
\psfrag{ gn}[t][t][0.8]{$GN$}
\psfrag{ lq}[t][t][0.8]{$LQT$}
\psfrag{GN}[t][t][0.8]{$GN$}
\psfrag{LQ}[t][t][0.8]{$LQT$}
\psfrag{dm}[t][t][0.9]{$d|\!\F\!|/dt$}
\psfrag{da}[t][t][0.9]{$d\angle\F/dt$}

\begin{subfigure}{1.\linewidth}    
\centering
\includegraphics[width=0.3\linewidth]{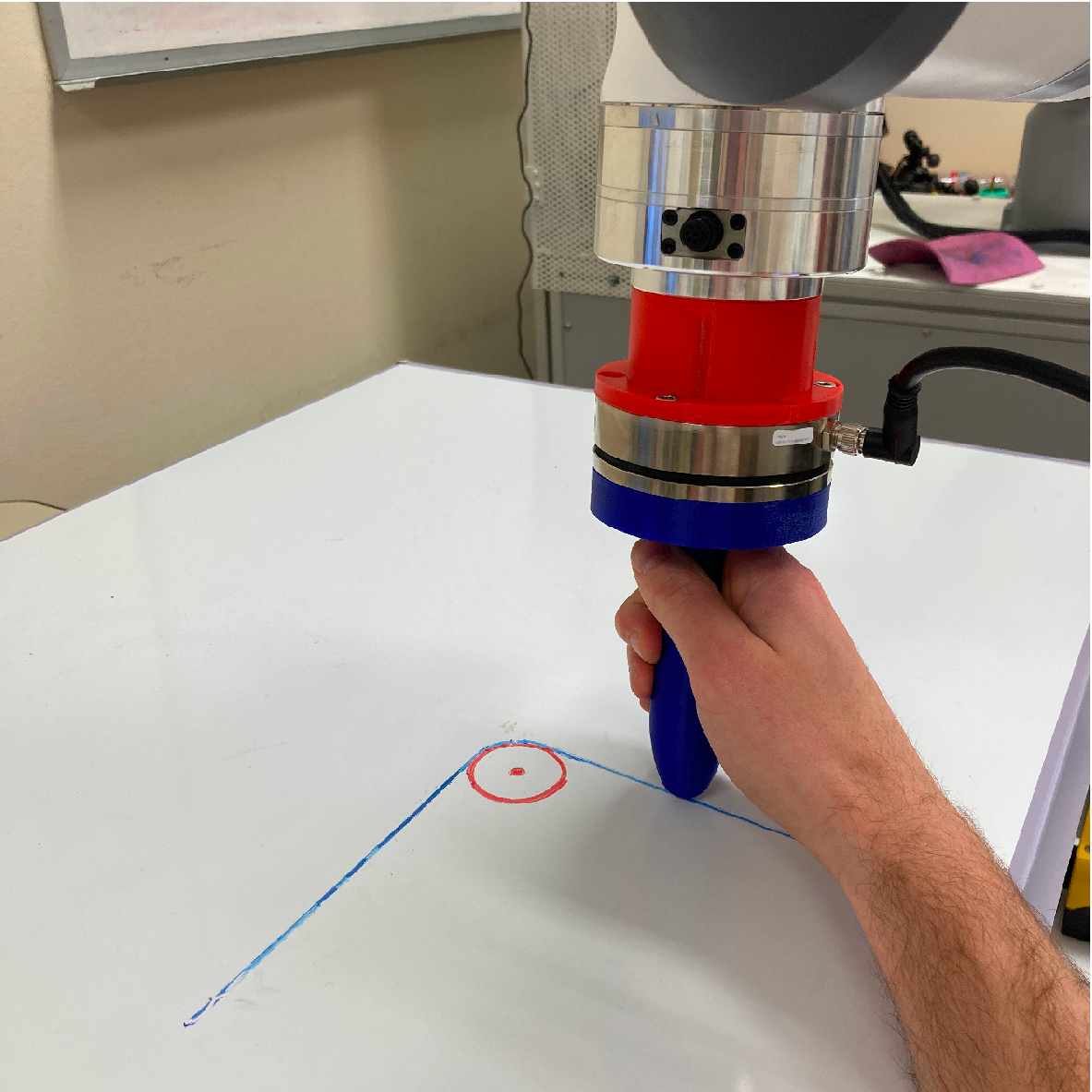}
\includegraphics[width=0.3\linewidth]{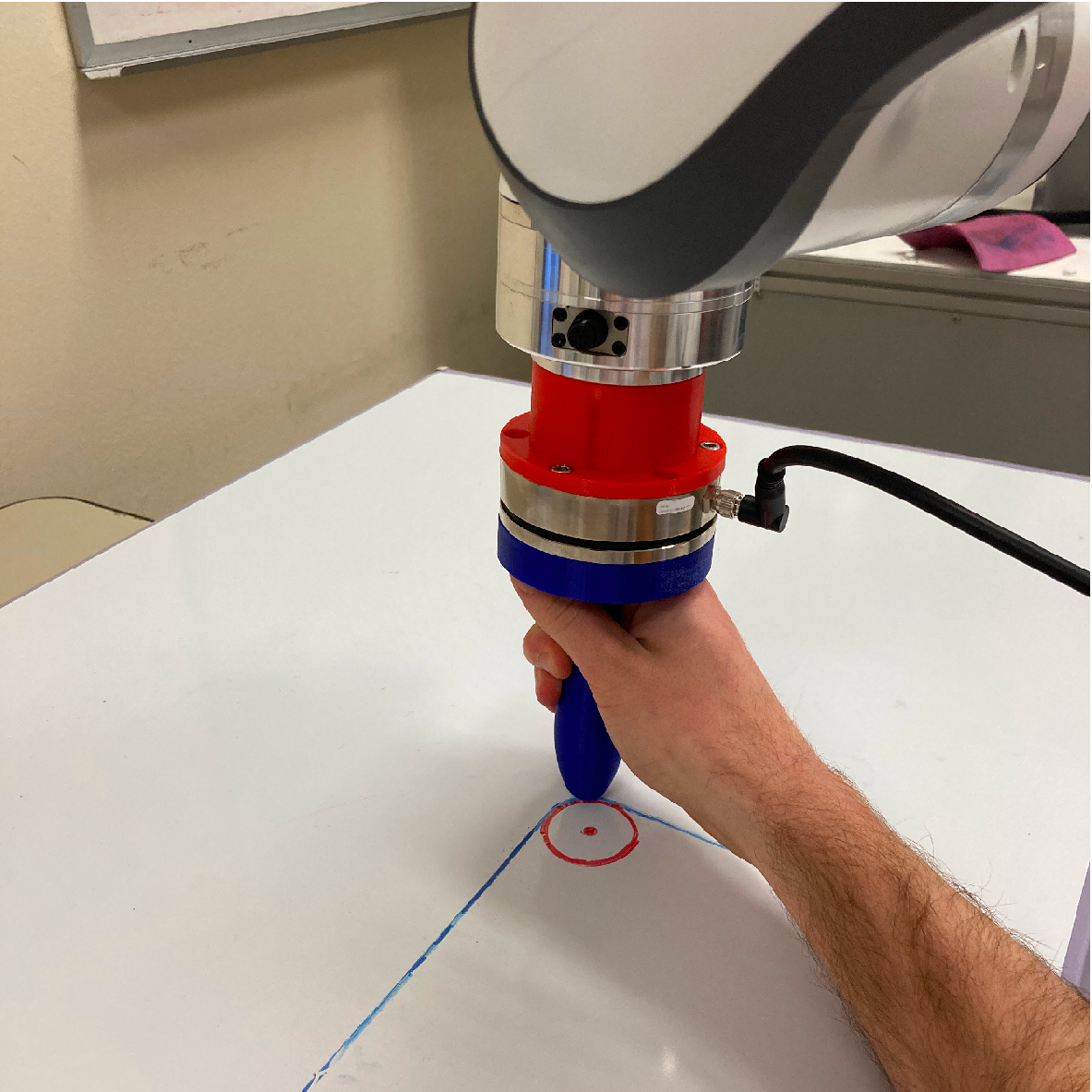}
\includegraphics[width=0.3\linewidth]{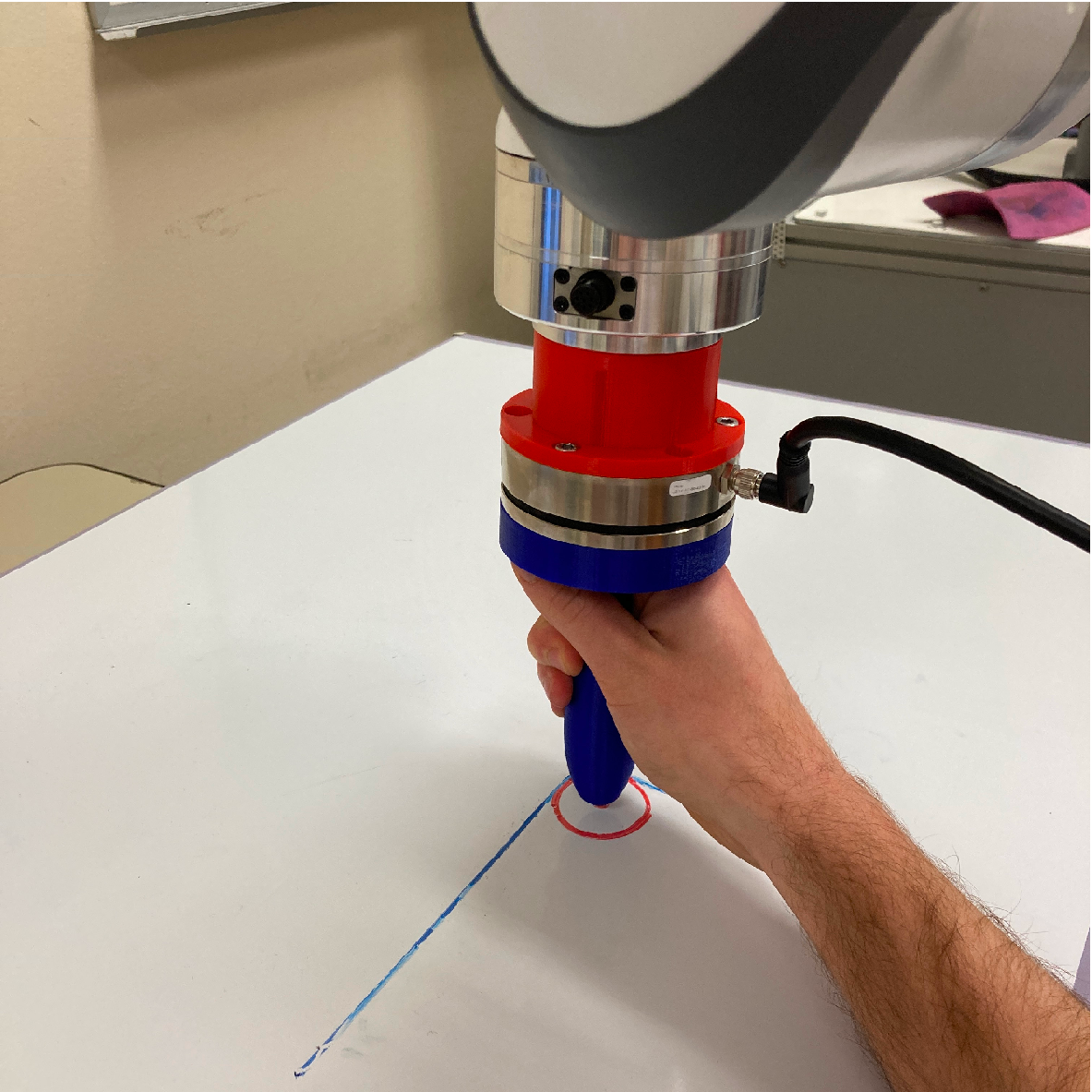}
    \caption{}
    \label{subfig:experiment1_snapshots}
\end{subfigure}\\[1mm]
\begin{subfigure}{0.55\linewidth}
\centering
    \includegraphics[width=\linewidth]{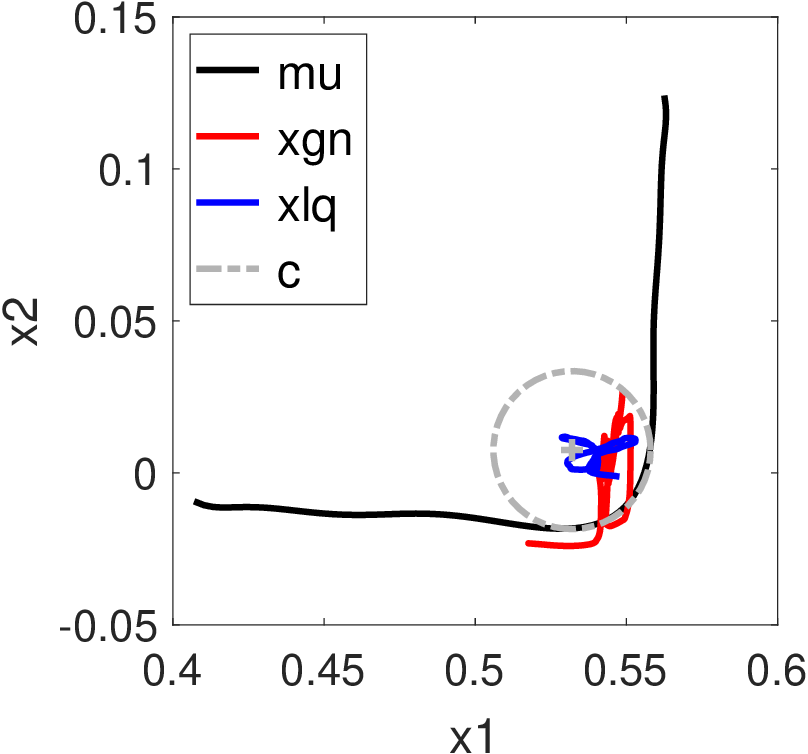}
    \caption{}
    \label{subfig:experiment1_results_1}
\end{subfigure}
\begin{subfigure}{0.9\linewidth}
\centering
    \includegraphics[width=\linewidth]{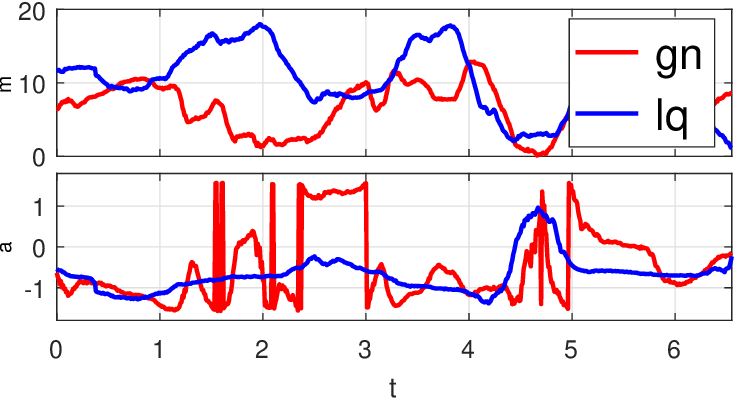} 
    \caption{}
    \label{subfig:experiment1_results_2}
\end{subfigure}\\[1mm]
\begin{subfigure}{0.45\linewidth}    
\centering
    \includegraphics[width=\linewidth]{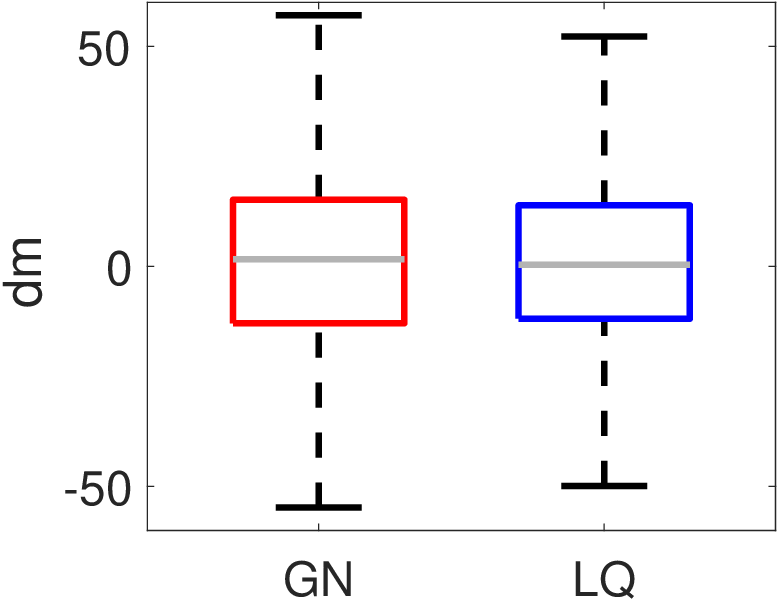} 
    \caption{}
    \label{subfig:experiment1_results_3}
\end{subfigure}
\begin{subfigure}{0.45\linewidth}
\centering
    \includegraphics[width=\linewidth]{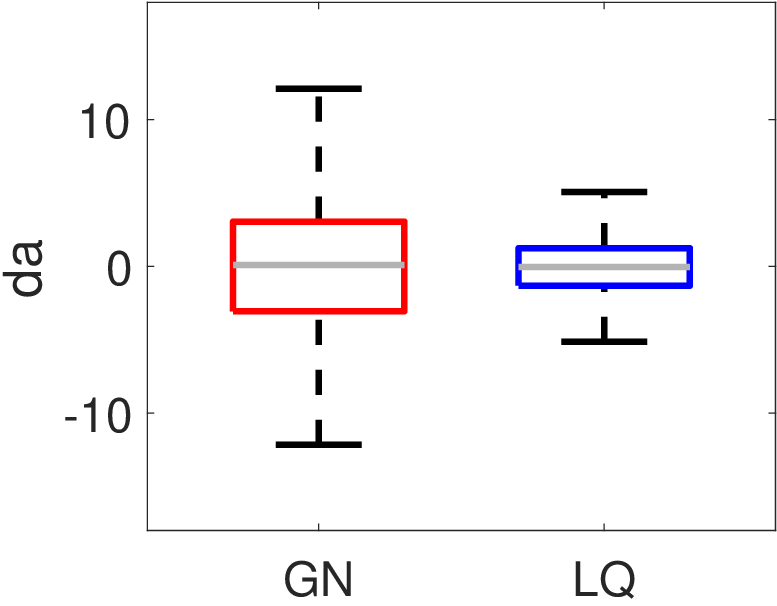} 
        \caption{}
        \label{subfig:experiment1_results_4}
\end{subfigure}
\caption{Experimental results of the center-reaching task. }
\label{fig:experiment1_results}
\vspace{-5mm}
\end{figure}

In Section~\ref{sec:methodology} we made some theoretical considerations about the behaviour of the GN algorithm close to EDS.
To validate so, an experiment consisting in moving the EE towards the center of the osculating circle $\boldsymbol{\gamma}(\hat{s})$ has been proposed. The performed task refers to Fig.~\ref{subfig:experiment1_snapshots}.

During the experiments three users were involved, two of them with no prior expertise in robotics. Each execution was recorded for approximately one minute. 
The numerical values chosen for the precision matrix $\Q$ were $\c_1 = 47.8 \cdot\!\boldsymbol{1}_3$, $\c_2=0.02\!\cdot\!\boldsymbol{1}_3$, $c_3=0.01$ and the control weight was set to $R=1e\!-\!5$ . 
The constraint curve $\m$ was obtained by kinesthetic demonstration over the curve of Fig.~\ref{subfig:experiment1_snapshots}.


In Fig.~\ref{fig:experiment1_results} the red and blue colors refer to the GN and LQT algorithms, respectively. 
Figure~\ref{subfig:experiment1_results_1} plots the recorded EE position $\x_{GN}$ and $\x_{LQT}$ extrapolated from an interval of $6s$; 
Fig.~\ref{subfig:experiment1_results_2} plots the variations of the force modulus $|\!\F\!|$ and its argument $\angle\F$; finally, Fig.~\ref{subfig:experiment1_results_3}-~\ref{subfig:experiment1_results_4} summarise the statistics of all the task executions in terms of the rate of change of the force modulus and argument, denoted with $d|\!\F\!|/dt$ and $d\angle\F/dt$.

The experiments underlined that the proximity to the center of the osculating circle $\boldsymbol{\gamma}(\hat{s})$ induces abrupt directional changes when using the GN algorithm. Precisely, Fig.~\ref{subfig:experiment1_results_1} demonstrates that reaching the center of $\boldsymbol{\gamma}(\hat{s})$ is not problematic in the LQT case, while it becomes complicated for the GN case. A major explanation of this phenomena is given in Fig.~\ref{subfig:experiment1_results_2}. Despite the modulus of the force $|\F|$ being in the same range for both GN and LQT, the former displays sharp variations for the argument $\angle \F$, induced by the proximity with an EDS.
This behaviour was observed throughout all the center-reaching task as revealed in the boxplots of Fig.~\ref{subfig:experiment1_results_3}-~\ref{subfig:experiment1_results_4}. 
On the left, one can observe that the rate of change of $|\F|$ is similar for both algorithms while, on the right, the GN case exhibits larger variations of the rate of change of $\angle\F$.

\subsection{Target following task}\label{subsec:target_following}
%
\begin{figure}
%
\psfrag{  aaa}[t][t][0.8]{$\m$}
\psfrag{  bbb}[t][t][0.8]{$\x_{GN}$}
\psfrag{  ccc}[t][t][0.8]{$\x_{LQT}$}
\psfrag{  ddd}[t][t][0.8]{$\x_{VM}$}
\psfrag{  fff}[t][t][0.8]{$\x_{GC}$}
\psfrag{x}[t][t][0.8]{$x$[m]}
\psfrag{y}[t][t][0.8]{$y$[m]}
\psfrag{z}[t][t][0.8]{$z$[m]}
\psfrag{t}[t][t][0.8]{$t$[s]}
\psfrag{gn}[][t][0.8]{$GN$}
\psfrag{lq}[][][0.8]{$LQT$}
\psfrag{ra}[][t][0.8]{$VM$}
\psfrag{ld}[][][0.8]{$GC$}
\psfrag{f}[t][t][1]{$||\F||$}
\psfrag{ft}[t][t][1]{$||\F||- |\F_{\tau}|$}
\psfrag{t}[t][t][1]{$t$[s]}
\psfrag{GN}[t][t][0.7]{$GN$}
\psfrag{ LQ}[t][t][0.7]{$LQT$}
\psfrag{ RA}[t][t][0.7]{$VM$}
\psfrag{s1}[t][t][1.2]{$s$}
\psfrag{s2}[t][t][1.2]{$\dot{s}$}
\psfrag{s3}[t][t][1.2]{$\ddot{s}$}
\centering
\begin{subfigure}{1.\linewidth}    
\centering
\includegraphics[width=0.3\linewidth]{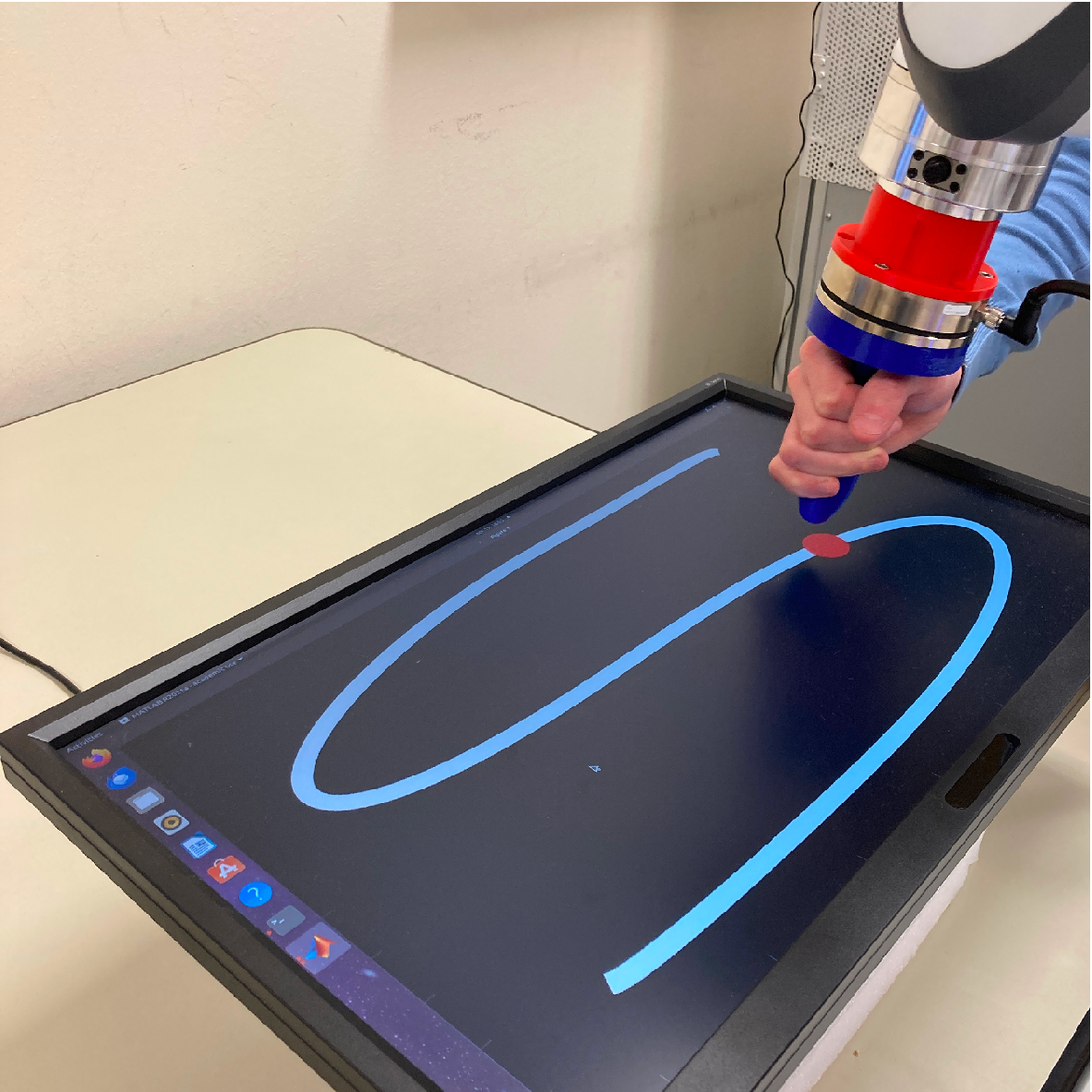}
\includegraphics[width=0.3\linewidth]{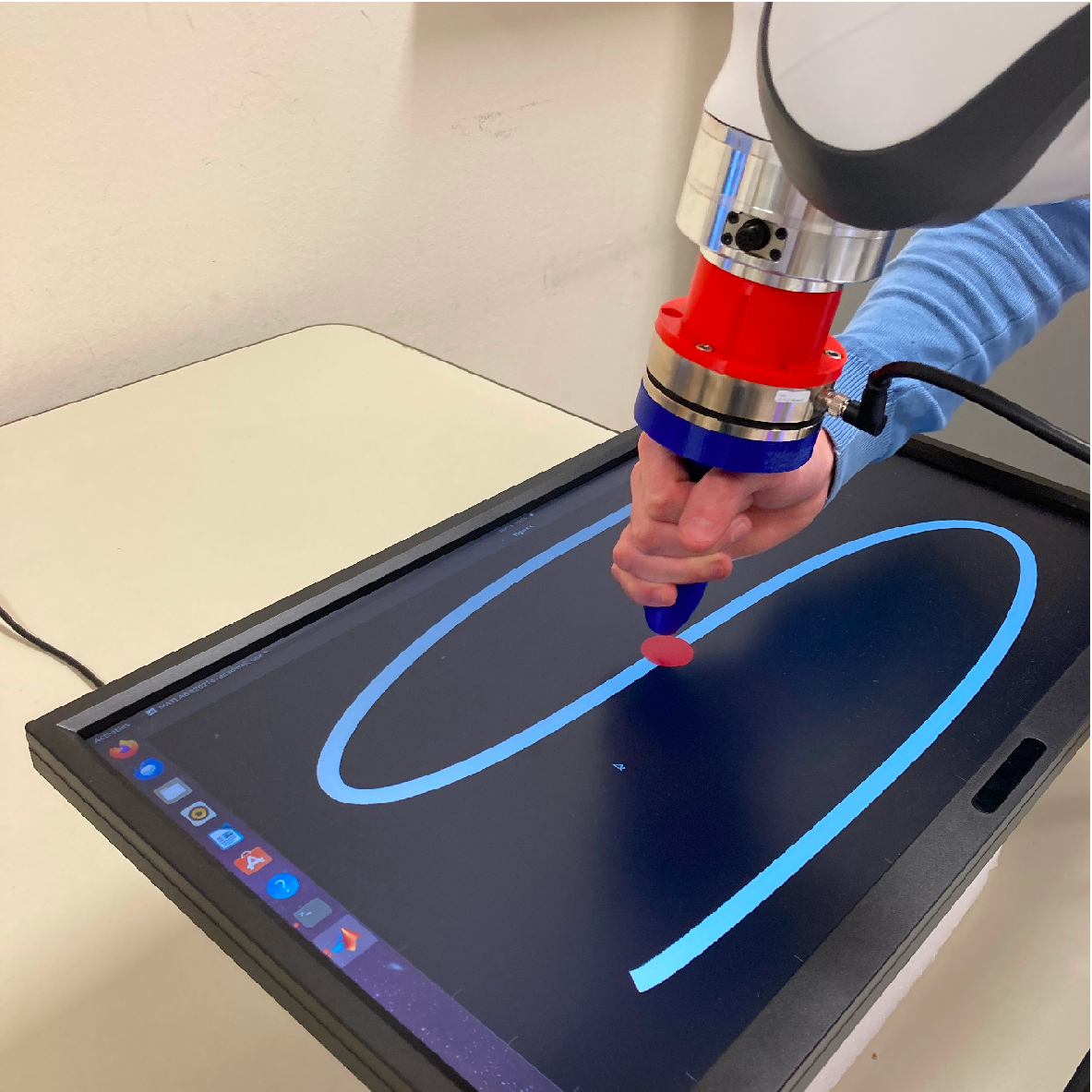}
\includegraphics[width=0.3\linewidth]{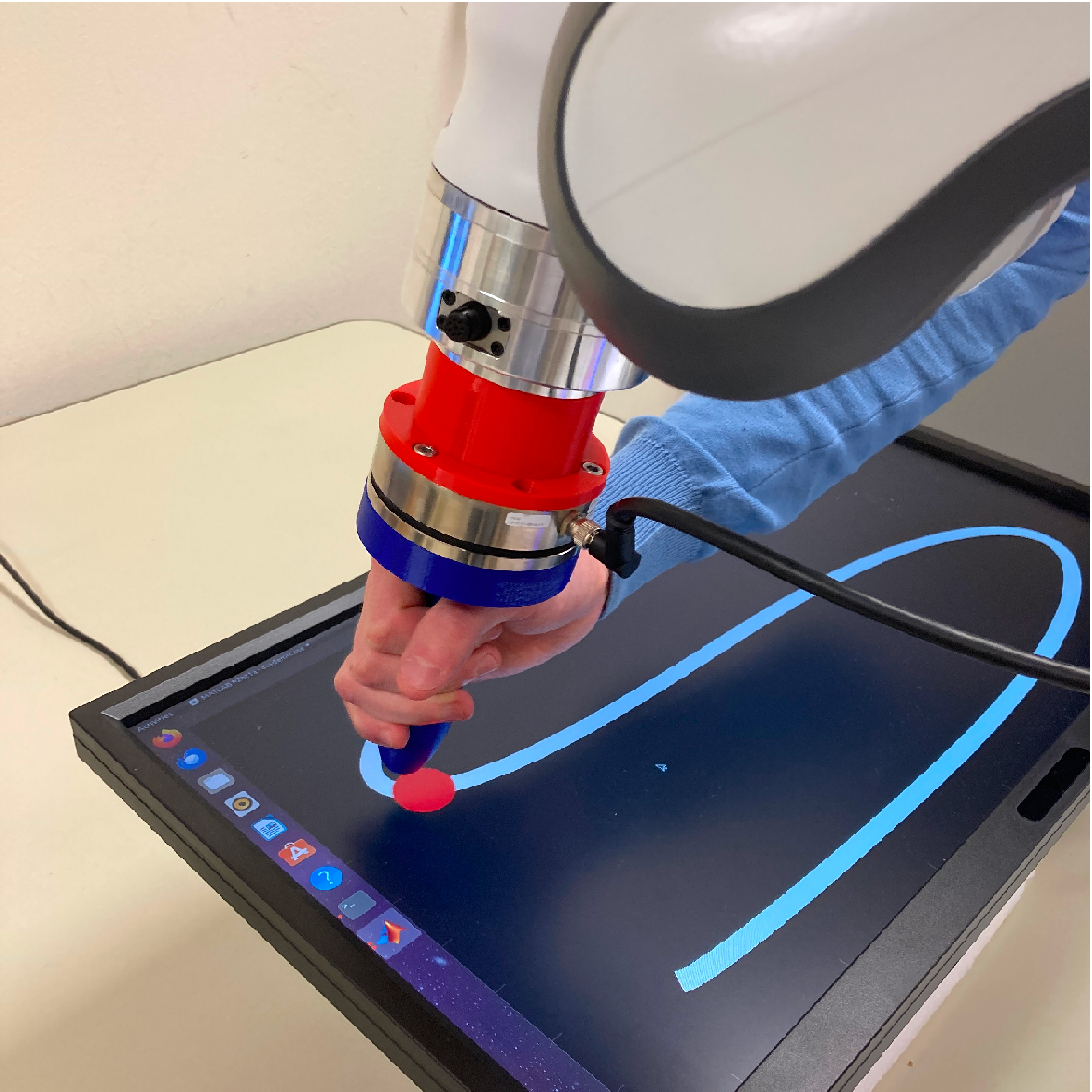}
    \caption{}
    \label{subfig:experiment2_snapshots}
\end{subfigure}\\[3mm]
\begin{subfigure}{0.9\linewidth}    
\centering
    \includegraphics[width=0.9\linewidth]{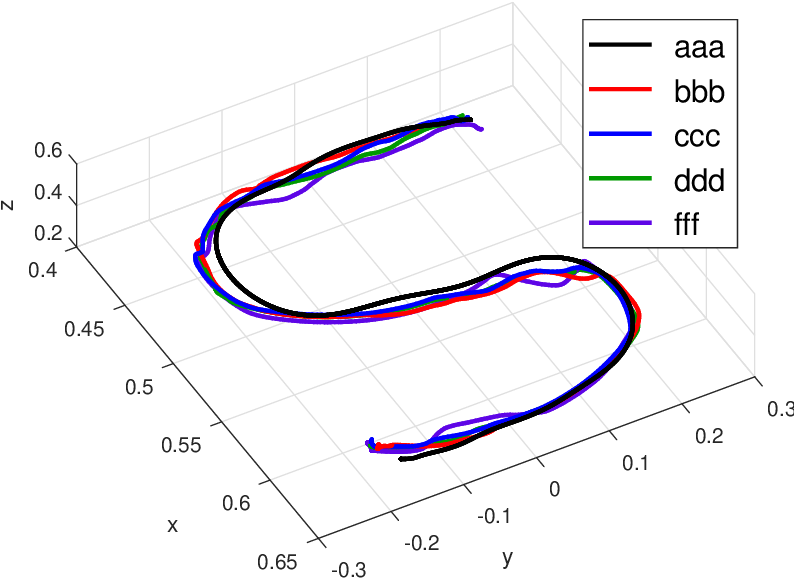} 
    \caption{}
    \label{subfig:experiment2_results_1}
\end{subfigure}\\[1mm]
\begin{subfigure}{0.45\linewidth}    
\centering
    \includegraphics[width=\linewidth]{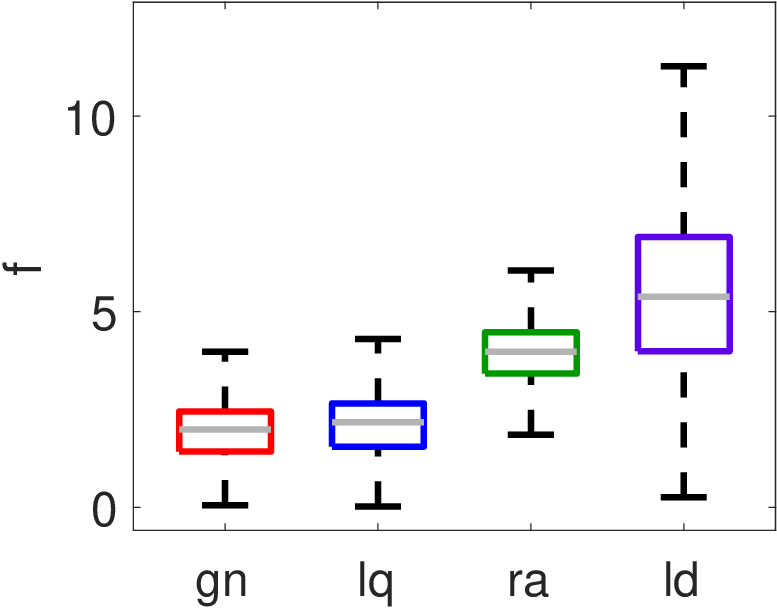} 
    \caption{}
    \label{subfig:experiment2_results_2}
\end{subfigure}
\begin{subfigure}{0.45\linewidth}
\centering
    \includegraphics[width=\linewidth]{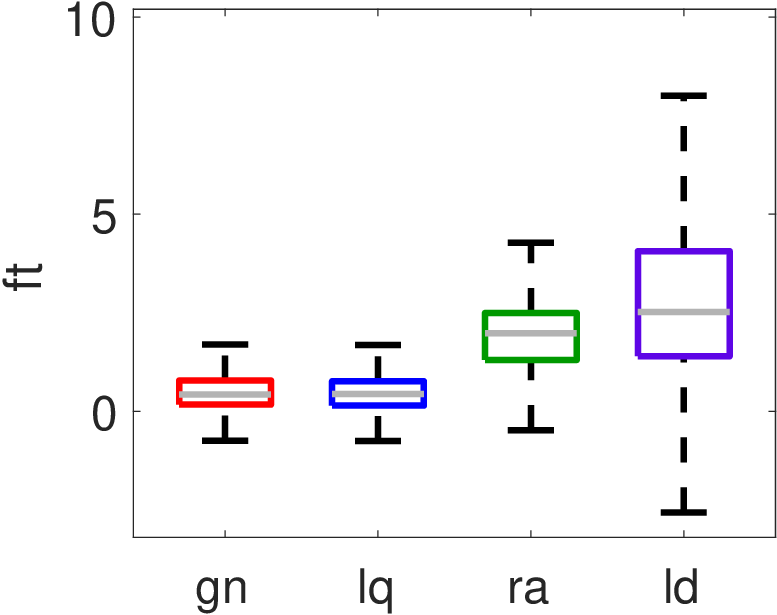} 
        \caption{}
        \label{subfig:experiment2_results_3}
\end{subfigure}\\[1mm]
\begin{subfigure}{0.9\linewidth}
\centering
    \includegraphics[trim=0cm 0cm 0cm 0.005cm,clip=true,width=\linewidth]{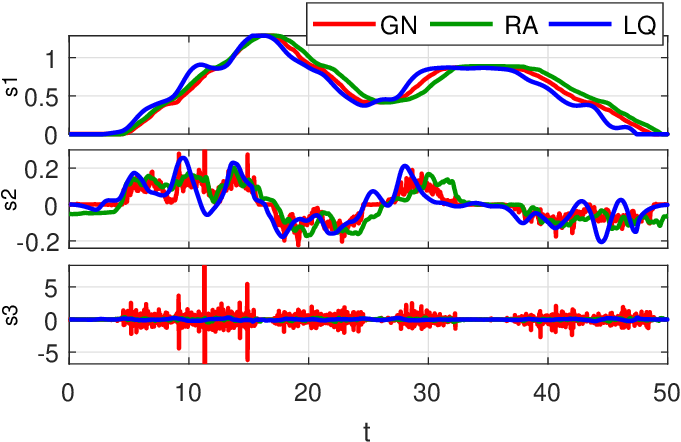} 
        \caption{}
        \label{subfig:experiment2_results_4}
\end{subfigure}
\caption{Experimental results of the target following task.} 
\label{fig:experiment2_results}
\end{figure}

In this experiment, the objective is to ensure that the EE's tip remains close to the moving point displayed on a second PC, as shown in Fig.~\ref{subfig:experiment2_snapshots}.
The aim is to prove that the LQT algorithm proposed in~\ref{algorithm1} is applicable to a real-time human-robot interaction scenario, and compare it with the GN, VM and GC performances. The VM methodology for phase calculation was introduced in~\cite{raiola2015co}.
The authors consider a virtual mechanism (VM) connected to the end effector of the robot and characterized by a spring-damper system, namely:
\begin{equation}\label{eq:vm_dyn_sys}
    \F_t = \K ( \x_t - \m(s_t) ) + \B ( \dot{\x}_t - \dot{\m}(s_t) ),
\end{equation}
where $\K$ and $\B$ are chosen to be the same as in~\eqref{eq:simulated_dyn}. 
The force exterted in the VM is always orthogonal to its velocity, that is $\m^{\prime}\!(s_t)^\top \F_t = 0$. By plugging~\eqref{eq:vm_dyn_sys} into this condition, one obtains:
\begin{equation}\label{eq:vm_ortho}
    \m^{\prime}\!(s_t)^\top \big( \K ( \x_t - \m(s_t) ) + \B ( \dot{\x}_t - \m^{\prime}\!(s_t) \dot{s}_t )  \big) = 0,
\end{equation}
from which the following update law holds
\begin{equation}\label{eq:vm_phase}
    \dot s_t \!=\! \big( \m^{\prime}\!(\!s_t\!)^\top \! \B \m^{\prime}\!(\!s_t\!) \big)^{\!-1} \! \m^{\prime}\!(\!s_t\!)^\top \! \big( \K \! ( \x_t - \m(\!s_t\!) ) + \B \dot \x_t  \big).
\end{equation}

During the experiments ten users were involved, four of them with no prior expertise in robotics. Every user had to execute the target following task eight times, alternating the right and left hand for each of the three tested algorithms. Each task execution lasted 50 seconds.
The numerical values chosen for the precision matrix $\Q$ were $\c_1 = 400.0\!\cdot\!\boldsymbol{1}_3$, $\c_2=0.14\!\cdot\!\boldsymbol{1}_3$, with $\boldsymbol{1}_3= [1,1,1]^\top$, $c_3=0.01$ and the control weight was set to $R=1e\!-\!5$. 
The constraint curve $\m$ was obtained by kinesthetic demonstration over the second PC monitor.

The results are summarized in Fig.~\ref{fig:experiment2_results}. Figure~\ref{subfig:experiment2_results_1} plots the recorded EE positions $\x = [ x_t, y_t, z_t ]^\top$ for the GN (red), LQT (blue), VM (green) and GC (violet) cases, while the constraint curve $\m = \m(s)$ is depicted in black. Given the target position $\m_t = [ \m_{p,t}, \m_{y,t}, \m_{z,t} ]^\top$, it is straightforward to see that in the four cases the robot succeeded to follow  $\m_t$ with a limited error $\e$.
Numerical values for the term $||\e|| = ||\x - \m(s)||$  are resumed in Table~\ref{tab:tab1} with their mean and standard deviation.

Interesting results come when analyzing the measurements of the forces applied to the end effector $\F = [ F_{x}, F_{y}, F_{z} ]^\top$.
It is reasonable to assume that the virtual constraint $\m$ simplifies the target following task, as it guides the robot EE along with point movements.
For the GC case, the user is not constrained to the curve $\m$, making it more difficult to perform the task.
The outcome, on average, is a significant larger effort required for the GC case, described in Fig.~\ref{subfig:experiment2_results_2} with the norm of the acquired force measurements. If we define $F_{\tau,t} = \m_t^\prime\cdot\F_t $ as the force component $\F_t$ projected to the tangential direction of the curve $\m_t^\prime = \partial \m(s_t)/\partial s_t$, the difference $||\F_t||- |\F_{\tau,t}|$ reported in Fig.~\ref{subfig:experiment2_results_3} quantifies the amount of force which is spent in holding the EE close to the curve $\m$. Again, as the GN, LQT and VM algorithms impose a virtual constrain, they exhibit substantial lower effort requirements if compared to the GC case, which makes the GC less suitable for this kind of application.

While GN, LQT and VM performances appear comparable, differences emerged when comparing the generated trajectories.
Indeed, the GN algorithm computes a velocity command with no smoothness cost on $\dot s$ and $\ddot s$, which affect the reference trajectories $\m(s_t)$, $\dot\m(s_t)$ and $\ddot\m(s_t)$ (see~\eqref{eq:velocity_acceleration}).
Conversely, the VM algorithm in~\eqref{eq:vm_phase} showed slight vibration, especially in regions with high curvature. This may be attributed to the choice of $\B$ and $\K$ parameters in~\eqref{eq:simulated_dyn}, which is critical for human-in-the-loop applications~\cite{ferraguti2019variable,braglia2024phase}.
This is exemplified by the experimental measurements reported in Fig.~\ref{subfig:experiment2_results_4}. Here we can appreciate that the LQT algorithm, by minimizing a jerk command, computes smoother derivatives $\dot s$ and $\ddot s$, demonstrating robustness against the difficulties encountered in the GN and VM cases. The smoothness has been quantitatively evaluated with the dimensionless squared jerk (DSJ)~\cite{chen2024variable}:
\begin{equation}\label{eq:dsj}
    DSJ(s) = \tau \sum_{t=t_0}^{T} \dddot s_t^2, \;\; DSJ(\dddot \x) = \tau \sum_{t=t_0}^{T}  ||\dddot \x_t||^2, 
\end{equation} 
with $\tau = (T-t_0)^5/L^2$. Parameters $T=50s$ and $L=3.5m$ represent the task duration and the path length, respectively. The metric in~\ref{eq:dsj} was applied both to the phase variable $s$ computed by the algorithms (absent for the GC case), and the actual end effector position $\x$. To reduce noise from computing derivatives, data for $DSJ(\dddot \x)$ were pre-processed using a moving average filter with a window size of $w=20$. The results in Table~\ref{tab:tab1} show an improvement in smoothness with the proposed LQT algorithm.

\begin{table}[t]
\centering
\caption{ Errors between EE and target position and DSJ.}\vspace{-4mm} 
\begin{center}
\begin{tabular}{|c|c|c|c|}
 \cline{2-4}
  \multicolumn{1}{c|}{}
& \multicolumn{1}{c|}{$||\e||$[cm]}
& \multicolumn{1}{c|}{DSJ($s$)}
& \multicolumn{1}{c|}{DSJ($\dddot \x$)}\\

\hline
\!GN\! & $ 2.5\!\pm\!1.16 $ & $ (\!1.02\!\pm\!3.446\!)e\!+\!14 $ & $ (\!5.44\!\pm\!5.570\!)e\!+\!16 $\\
\hline
\!LQT\! & $ 2.3\!\pm\!1.31 $ & $ \boldsymbol{(\!3.61\!\pm\!5.771\!)e\!+\!9} $ & $ \boldsymbol{(\!7.40\!\pm\!0.071\!)e\!+\!15} $ \\
\hline
\!VM\! & $ 2.3\!\pm\!1.26 $  & $ (\!1.22\!\pm\!3.446\!)e\!+\!11 $ & $ (\!1.53\!\pm\!1.129\!)e\!+\!16 $ \\
\hline
\!GC\! & $ 2.2\!\pm\!1.14 $  & $ \times $  & $ (\!1.18\!\pm\!1.347\!)e\!+\!17 $ \\
\hline
\end{tabular}
\label{tab:tab1}
\end{center}
\vspace{-4mm}
\end{table}

\subsection{Discussion}\label{subsec:discussion}


Results from the target following task confirm the previous statement, with higher force demand when no virtual constraint is implemented. Indeed, this increases cognitive stress for the user, who must track the target point while ensuring the EE follows the specified curve.
Results also highlights that incorporating the proposed LQT framework has the advantage, over a GN and VM implementations, of generating smoother trajectories as evinced by the results in Fig.~\ref{subfig:experiment2_results_4} and Table~\ref{tab:tab1}.
By specifying the precision matrix $\Q$ and the control weight $R$, one can robustly contain unwanted vibrations by minimizing a jerk command and guaranteed a smooth execution of the proposed guiding tasks~\cite{todorov1998smoothness,sharkawy2021minimum}. 


Nevertheless, a high cost on the jerk command can reduce the algorithm tracking precision, as it penalizes high acceleration profiles~\cite{meirovitch2016geometrical}. In the target following task, this effect could alter the perceived inertia of the robot when moving along the task curve. Although lower accelerations may impact the robot's responsiveness~\cite{palleschi2019time}, we demonstrate in the center-reaching task the importance of imposing a smoothness cost. Precisely, when reaching EDSs, the solution of the optimization cost in~\eqref{eq:optimal_problem} degenerates~\cite{luenberger1984linear}. Using the GN algorithm, experiments underlined the presence of sharp changes in the force direction exerted to the EE. This is caused by the generation of abrupt velocity commands in~\eqref{eq:delta_s}, as opposed to the proposed algorithm which, instead, successfully prevents this condition.

\section{Conclusion}\label{sec:conclusion}

We proposed an approach based on a linear quadratic tracking (LQT) algorithm to regulate the position of the end effector (EE) of the robot along a virtual constraint. Our approach allows to generate smooth trajectories, avoiding abrupt changes in the derivatives of the EE's reference position. Results demonstrate that our approach is robust against the proximity to Euclidean Distance Singularities (EDS), which is not the case for a solution based on a Gauss-Newton (GN) framework.
In the experimental evaluation, the proposed LQT algorithm was compared with GN, a virtual mechanism (VM), and a gravity compensation (GC) guidance, displaying higher reliability in the LQT case. In future works, we plan to extend our methodology to a user study, incorporating subject-related metrics to better evaluate the user perspective.

\bibliographystyle{IEEEtran}

\bibliography{references} 

\begin{thebibliography}{10}
\providecommand{\url}[1]{#1}
\csname url@samestyle\endcsname
\providecommand{\newblock}{\relax}
\providecommand{\bibinfo}[2]{#2}
\providecommand{\BIBentrySTDinterwordspacing}{\spaceskip=0pt\relax}
\providecommand{\BIBentryALTinterwordstretchfactor}{4}
\providecommand{\BIBentryALTinterwordspacing}{\spaceskip=\fontdimen2\font plus
\BIBentryALTinterwordstretchfactor\fontdimen3\font minus \fontdimen4\font\relax}
\providecommand{\BIBforeignlanguage}[2]{{%
\expandafter\ifx\csname l@#1\endcsname\relax
\typeout{** WARNING: IEEEtran.bst: No hyphenation pattern has been}%
\typeout{** loaded for the language `#1'. Using the pattern for}%
\typeout{** the default language instead.}%
\else
\language=\csname l@#1\endcsname
\fi
#2}}
\providecommand{\BIBdecl}{\relax}
\BIBdecl

\bibitem{rosenberg1993use}
L.~B. Rosenberg \emph{et~al.}, ``The use of virtual fixtures to enhance operator performance in time delayed teleoperation,'' \emph{Journal of Dynamic Systems Control}, vol.~49, pp. 29--36, 1993.

\bibitem{selvaggio2018passive}
M.~Selvaggio, G.~A. Fontanelli, F.~Ficuciello, L.~Villani, and B.~Siciliano, ``Passive virtual fixtures adaptation in minimally invasive robotic surgery,'' \emph{IEEE Robot. Automat. Lett.}, vol.~3, pp. 3129--3136, 2018.

\bibitem{bowyer2013active}
S.~A. Bowyer, B.~L. Davies, and F.~R. y~Baena, ``Active constraints/virtual fixtures: A survey,'' \emph{IEEE Trans. Robot.}, vol.~30, pp. 138--157, 2013.

\bibitem{raiola2015co}
G.~Raiola, X.~Lamy, and F.~Stulp, ``Co-manipulation with multiple probabilistic virtual guides,'' in \emph{2015 IEEE/RSJ IROS}, 2015, pp. 7--13.

\bibitem{amirabdollahian2002minimum}
F.~Amirabdollahian, R.~Loureiro, and W.~Harwin, ``Minimum jerk trajectory control for rehabilitation and haptic applications,'' in \emph{Proceedings 2002 IEEE ICRA}, vol.~4, 2002, pp. 3380--3385.

\bibitem{sciavicco}
L.~Sciavicco, B.~Siciliano, L.~Villani, and G.~Oriolo, \emph{Robotics: Modelling, planning and Control, ser. Advanced Textbooks in Control and Signal Processing}.\hskip 1em plus 0.5em minus 0.4em\relax Berlin, Germany: Springer-Verlag, 2011.

\bibitem{verscheure2009time}
D.~Verscheure, B.~Demeulenaere, J.~Swevers, J.~De~Schutter, and M.~Diehl, ``Time-optimal path tracking for robots: A convex optimization approach,'' \emph{IEEE Trans. on Automatic Control}, vol.~54, pp. 2318--2327, 2009.

\bibitem{bianco2017scaling}
C.~G.~L. Bianco and F.~Ghilardelli, ``A scaling algorithm for the generation of jerk-limited trajectories in the operational space,'' \emph{Robotics and Computer-Integrated Manufacturing}, vol.~44, pp. 284--295, 2017.

\bibitem{braglia2023online}
G.~Braglia, M.~Tagliavini, F.~Pini, and L.~Biagiotti, ``Online motion planning for safe human–robot cooperation using b-splines and hidden markov models,'' \emph{Robotics}, vol.~12, 2023.

\bibitem{ijspeert2013dynamical}
A.~J. Ijspeert, J.~Nakanishi, H.~Hoffmann, P.~Pastor, and S.~Schaal, ``Dynamical movement primitives: learning attractor models for motor behaviors,'' \emph{Neural computation}, vol.~25, pp. 328--373, 2013.

\bibitem{ijspeert2002movement}
A.~J. Ijspeert, J.~Nakanishi, and S.~Schaal, ``Movement imitation with nonlinear dynamical systems in humanoid robots,'' in \emph{Proceedings 2002 IEEE ICRA}, vol.~2, 2002, pp. 1398--1403.

\bibitem{sidiropoulos2021reversible}
A.~Sidiropoulos and Z.~Doulgeri, ``A reversible dynamic movement primitive formulation,'' in \emph{2021 IEEE ICRA}, 2021, pp. 3147--3153.

\bibitem{braglia2024phase}
G.~Braglia, D.~Tebaldi, and L.~Biagiotti, ``Phase-free dynamic movement primitives applied to kinesthetic guidance in robotic co-manipulation tasks,'' \emph{arXiv:2401.08238}, 2024.

\bibitem{reynoso2013time}
P.~Reynoso-Mora, W.~Chen, and M.~Tomizuka, ``On the time-optimal trajectory planning and control of robotic manipulators along predefined paths,'' in \emph{2013 American Control Conference}, 2013, pp. 371--377.

\bibitem{girgin2023projection}
H.~Girgin, T.~L{\"o}w, T.~Xue, and S.~Calinon, ``Projection-based first-order constrained optimization solver for robotics,'' \emph{arXiv:2306.17611}, 2023.

\bibitem{bischof2016combined}
B.~Bischof, T.~Gl{\"u}ck, and A.~Kugi, ``Combined path following and compliance control for fully actuated rigid body systems in 3-d space,'' \emph{IEEE Trans. on Control Systems Technology}, vol.~25, pp. 1750--1760, 2016.

\bibitem{zhang2020assist}
L.~Zhang, S.~Guo, and Q.~Sun, ``An assist-as-needed controller for passive, assistant, active, and resistive robot-aided rehabilitation training of the upper extremity,'' \emph{Applied Sciences}, vol.~11, p. 340, 2020.

\bibitem{wang2012gauss}
Y.~Wang, ``Gauss--newton method,'' \emph{Wiley Interdisciplinary Reviews: Computational Statistics}, vol.~4, pp. 415--420, 2012.

\bibitem{khatib2022constraint}
O.~Khatib, M.~Jorda, J.~Park, L.~Sentis, and S.-Y. Chung, ``Constraint-consistent task-oriented whole-body robot formulation: Task, posture, constraints, multiple contacts, and balance,'' \emph{The International Journal of Robotics Research}, vol.~41, pp. 1079--1098, 2022.

\bibitem{rcfs}
``Robotics codes from scratch ({RCFS}),'' \url{https://rcfs.ch/}, Accessed: 2024.

\bibitem{luenberger1984linear}
D.~G. Luenberger, Y.~Ye \emph{et~al.}, \emph{Linear and nonlinear programming}.\hskip 1em plus 0.5em minus 0.4em\relax Springer, 1984, vol.~2.

\bibitem{palleschi2019time}
A.~Palleschi, M.~Garabini, D.~Caporale, and L.~Pallottino, ``Time-optimal path tracking for jerk controlled robots,'' \emph{IEEE Robot. Automat. Lett.}, vol.~4, pp. 3932--3939, 2019.

\bibitem{piazzi2000global}
A.~Piazzi and A.~Visioli, ``Global minimum-jerk trajectory planning of robot manipulators,'' \emph{IEEE Trans. on industrial electronics}, vol.~47, pp. 140--149, 2000.

\bibitem{biess2007computational}
A.~Biess, D.~G. Liebermann, and T.~Flash, ``A computational model for redundant human three-dimensional pointing movements: integration of independent spatial and temporal motor plans simplifies movement dynamics,'' \emph{Journal of Neuroscience}, vol.~27, pp. 13\,045--13\,064, 2007.

\bibitem{sharkawy2021minimum}
A.-N. Sharkawy, ``Minimum jerk trajectory generation for straight and curved movements: Mathematical analysis,'' \emph{arXiv:2102.07459}, 2021.

\bibitem{todorov1998smoothness}
E.~Todorov and M.~I. Jordan, ``Smoothness maximization along a predefined path accurately predicts the speed profiles of complex arm movements,'' \emph{Journal of Neurophysiology}, vol.~80, pp. 696--714, 1998.

\bibitem{calinon2020mixture}
S.~Calinon, ``Mixture models for the analysis, edition, and synthesis of continuous time series,'' \emph{Mixture Models and Applications}, pp. 39--57, 2020.

\bibitem{biagiotti2008trajectory}
L.~Biagiotti and C.~Melchiorri, \emph{Trajectory planning for automatic machines and robots}.\hskip 1em plus 0.5em minus 0.4em\relax Springer Science \& Business Media, 2008.

\bibitem{toponogov2006differential}
V.~A. Toponogov, \emph{Differential geometry of curves and surfaces}.\hskip 1em plus 0.5em minus 0.4em\relax Springer, 2006.

\bibitem{chen2024variable}
H.~Chen, W.~Xu, W.~Guo, and X.~Sheng, ``Variable admittance control using velocity-curvature patterns to enhance physical human-robot interaction,'' \emph{IEEE Robot. Automat. Lett.}, 2024.

\bibitem{meirovitch2016geometrical}
Y.~Meirovitch, D.~Bennequin, and T.~Flash, ``Geometrical invariance and smoothness maximization for task-space movement generation,'' \emph{IEEE Trans. Robot.}, vol.~32, pp. 837--853, 2016.

\bibitem{priess2014solutions}
M.~C. Priess, R.~Conway, J.~Choi, J.~M. Popovich, and C.~Radcliffe, ``Solutions to the inverse lqr problem with application to biological systems analysis,'' \emph{IEEE Trans. on Control Systems Technology}, vol.~23, pp. 770--777, 2014.

\bibitem{ferraguti2019variable}
F.~Ferraguti, C.~Talignani~Landi, L.~Sabattini, M.~Bonfe, C.~Fantuzzi, and C.~Secchi, ``A variable admittance control strategy for stable physical human--robot interaction,'' \emph{The International Journal of Robotics Research}, vol.~38, pp. 747--765, 2019.

\end{thebibliography}

\end{document}